# Instant Learning: Parallel Deep Neural Networks and Convolutional Bootstrapping

Andrew J.R. Simpson [#1]

[#] Centre for vision, speech and signal processing (CVSSP), University of Surrey,
Guildford, Surrey, UK
[1] `Andrew.Simpson@surrey.ac.uk`

*Abstract*—**Although deep neural networks (DNN) are able to scale with direct advances in computational power (e.g., memory and processing speed), they are not well suited to exploit the recent trends for parallel architectures. In particular, gradient descent is a sequential process and the resulting serial dependencies mean that DNN training cannot be parallelized effectively. Here, we show that a DNN may be replicated over a massive parallel architecture and used to provide a cumulative sampling of local solution space which results in rapid and robust learning. We introduce a complimentary convolutional bootstrapping approach that enhances performance of the parallel architecture further. Our parallelized convolutional bootstrapping DNN out-performs an identical fully-trained traditional DNN after only a single iteration of training.**

I. INTRODUCTION

Deep neural networks (DNN) are typically trained by a process of back-propagated gradient descent [1]-[4]. First, random starting points are assigned to the weights of the network, then an error function is used to compute gradients which can be inverted to provide local directions towards minimizing error. DNNs present a non-convex optimization problem and hence the ultimate *local solution* depends upon the random starting weights [5]. Therefore, the process of training a single instance of a DNN using gradient descent may be interpreted as sampling (once) the local solution space.

It has been demonstrated that the various local solutions, that may be sampled in this way, are typically equivalent [5]. However, they are not the same and hence the predictions of the respective models may not be the same. Furthermore, if each independently sampled local solution features some degree of *local overfitting*, then the overfitting in each sampled model may be somewhat exclusive (local). Hence, in principle, by taking a wide sample of the local solution space we obtain a convergent set of predictions and a diffuse set of local overfitting characteristics. If we then integrate the predictions, we may expect to enhance the mutual (correct) predictions whilst suppressing the non-mutual errors (due to local overfitting). Extending this idea further, some means to make each local instance of the training data independent (e.g., such as 'data augmentation' [4]) would lead to further enhancements via the cumulative sampling process.

In this paper, we demonstrate that this parallel local sampling approach can be used to provide both 'instantaneous' learning and performance gains capable of exploiting the recent trend for parallel computing architectures. Within the context of the well-known MNIST [3] hand-written digit classification problem, we replicate Hinton's original MNIST DNN [1] a large number of times and train each from different starting weights. We then integrate the predictions of each model and show that, cumulatively, this results in improved performance as compared to a single instance of the DNN. Then, to address the question of data augmentation in a bootstrapping sense (i.e., without prior knowledge beyond the training data), we introduce a convolutional bootstrapping procedure. For each local model, we select (at random) a single training image and we convolve the entire local training and test set with that image. This ensures that each local model encounters a local feature space that is independent but within the global feature space, and hence enhances the independence of local learning and local overfitting. We illustrate the strengths of this parallel approach by computing test classification error as a function of training iterations and as a function of the scale of the architecture.

II. METHOD

To illustrate the complimentary ideas of parallel local sampling and convolutional bootstrapping, we chose the well-known computer vision problem of hand-written digit classification using the MNIST dataset [3], [1], [4]. For the input layer we unpacked the images of 28x28 pixels into vectors of length 784. An example digit is given in Fig. 1. Pixel intensities were normalized to zero mean. Replicating

Hinton's [1] architecture, but using the biased sigmoid activation function [6], we built a fully connected network of size 784x100x10 units, with a 10-unit softmax output layer, corresponding to the 10-way digit classification problem. This represented a single *n*th unit (instance) of the (*N* sized) parallel architecture.

*Local models*. Each instance of the model, which we call a local model, was independently trained using stochastic gradient descent (SGD) from different random starting weights. We trained the various instances of the local model on the same 60,000 training examples from the MNIST dataset [4]. Each iteration of SGD consisted of a complete sweep of the entire training data set. The resulting local models were then used as parallel feed-forward devices to process the 10,000 separate test examples at various (full-sweep) iteration points of training. After the entire pool of local models were trained and applied to the test data, the output layers were accumulated in an additive element-wise fashion (without application of the softmax stage) before being averaged. All models were trained without dropout.

The resulting integrated predictions (for the entire test set) were input to the softmax function to produce the final predictions. Finally, the predictions generated using the parallel architecture were tested against the classifications of the test data labels and an error rate computed. This procedure was replicated for *N* local models and for various numbers of training iterations (per local model).

*Convolutional bootstrapping*. We also replicated the above parallel sampling procedure with an additional bootstrap-styled 'data augmentation' stage. For each local model, prior to training, a single 'local feature space' training image was selected at random and convolved with the entire set of training and test images (for that local instance of the model). This local convolution involved element-wise multiplication of each training and test image with the selected local-feature-space image. Different local-feature-space images were selected (at random) for each local model. The local model was then trained and tested on these locally convolved training and test sets.

For comparison to the two respective parallel architectures, we also trained a single instance of the same (local) model for 1000 equivalent iterations (full-sweep, SGD) using the same training data (without convolutional bootstrap) and using identical SGD parameters. This model, representing the traditional approach, was then tested on the same test data. Hence, the training data and parameterization were held constant (though the effective parameter cost of the parallel architectures was cumulatively far larger). Thus, we could compare the serial and parallel architectures at equivalent stages of training and also contrast the efficacy of iterations used in parallel versus training iterations used in series.

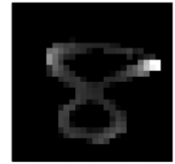

**Fig. 1. Example MNIST image.** We took the 28x28 pixel images and unpacked them into a vector of length 784 to form the input at the first layer of the local model (DNN).

III. RESULTS

Fig. 2a plots the classification error as a function of SGD training iterations for a single instance of the unit model. This provides a benchmark to which we may compare the parallel versions of the same architecture. We will refer to this model as the 'traditional' model from here onwards. Performance saturates after around 100 iterations of training. Fig. 2b plots classification error as a function of the number (*N*) of local models used in parallel for various numbers of iterations of training (without convolutional bootstrapping). After around $N \geq 1000$ local models, even for the parallel architecture trained for only a single iteration, the integrated cumulative prediction accuracy approximately equals that of the fully trained traditional model. This advantage increases with the number of parallel training iterations and with the number of local models. At 6 parallel training iterations, and for $N >> \sim 1000$ local models, the classification error rate is reduced by around 30% compared to the fully trained traditional model. In fact, even on a cumulative per-iteration basis, performance improves more rapidly for the parallel architectures. In other words, if 100 iterations of training are to be spent, regardless of temporal dependency constraints (i.e., in either serial or parallel), they are best spent on the parallel architecture with 6 iterations of training per local model.

Fig. 2c plots the same for the convolutional bootstrapping version of the parallel architecture. In this case, performance even for the single iteration of parallel training exceeds the best performance of the fully trained traditional model. When several parallel training iterations are used performance is enhanced considerably beyond that shown for the parallel architecture without convolutional bootstrapping. At 6 parallel training iterations, the classification error rate is reduced by around 50% compared with the fully trained traditional model.

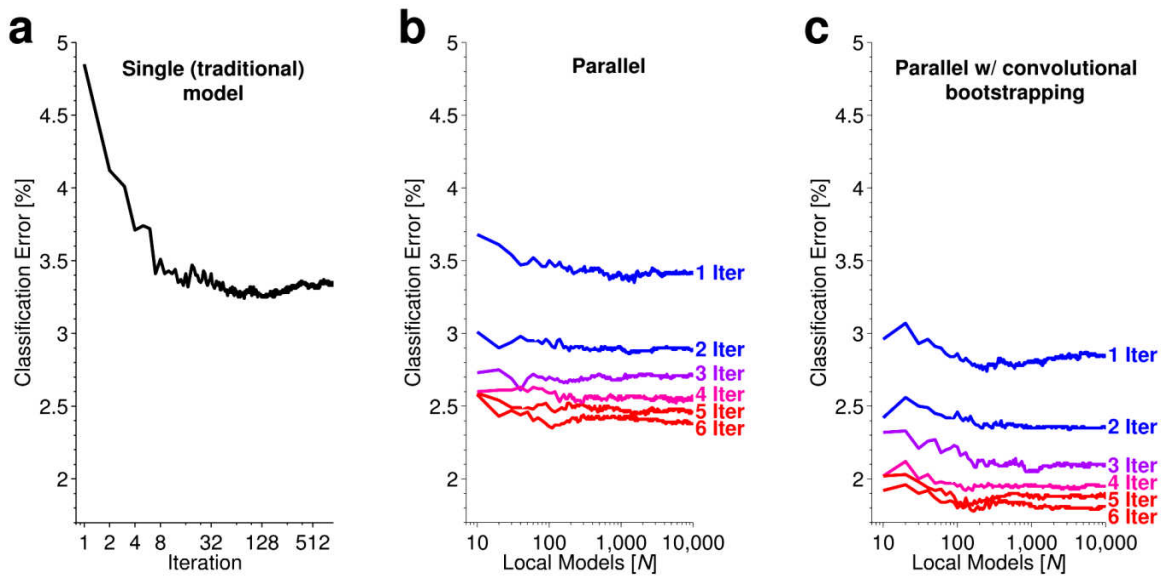

**Fig. 2. Serial Versus Parallel and Convolutional Bootstrapping. a** Test classification error as a function of training iterations (SGD) for a 'traditional DNN' **b** Test classification error as a function of the number of local models (*N*) for the parallel architecture (featuring the same local network architecture and training regime as the 'traditional model'). **c** Test classification error as a function of the number of local models (*N*) for the parallel architecture featuring convolutional bootstrapping. Note: iterations ('Iter' on the above plots) indicate a full sweep (SGD) of training data.

## IV. DISCUSSION AND CONCLUSIONS

We have demonstrated that the DNN solution space may be sampled using local models in parallel and the local predictions accumulated to provide both improved learning rate and improved performance. In particular, we have demonstrated that a parallel architecture, featuring multiple local models, is able to match (and even surpass) the performance of a fully-trained traditional (single) model after only a single (parallel) iteration of training. Thus, in principle, given suitably scaled parallel computational resources, a given model might be trained 'instantaneously' by similar parallelization.

We have also introduced a novel convolutional bootstrapping technique designed to increase the degree of independence between the local solutions of the parallel local models. The intuition behind this procedure is that, by convolution with single exemplars of the training data, each local model is then constrained to a local feature space and hence is 'corralled' into fitting and overfitting within the unique context of that feature space. We have demonstrated that this convolutional bootstrapping allows parallel performance to be further enhanced. This is a form of 'data augmentation' [4]. However, unlike data augmentation approaches featuring arbitrary transforms [4] or assumptions about the representation of the data, the described convolutional bootstrapping can be applied without prior knowledge or assumptions about the nature and/or structure of the data. Furthermore, unlike conventional ideas of data augmentation, we do not use the local data as additional constraints on a single model but rather as independent constraints on independent models. Hence, to some extent we are not trying to regularize to avoid overfitting but rather to channel or map our overfitting into complimentary local feature spaces. It remains to be seen how techniques such as dropout [7], [8] and interpretations from sampling theory [9] might be related or integrated.

The obvious advantages of the parallel sampling approach described here include applicability to industrial machine learning problems and scalability – the accuracy of the sampling and cumulative prediction scales arbitrarily. Furthermore, the approach provides a robust distributed architecture where individual local models may be 'lost' without much impact on the cumulative result. Hence, parallel learning and feed-forward prediction are robust given suitably large (and distributed) architectures. This means that the approach may be suitable for online (i.e., network connected) learning problems. Furthermore, issues relating to over-training (and overfitting) may be partially or wholly mitigated in the parallel architecture which is capable of learning with very few iterations of training.

ACKNOWLEDGMENT

AJRS did this work on the weekends and was supported by his wife and children.